\documentclass[12pt]{article}

\usepackage{sbc-template}
\usepackage{graphicx}
\usepackage{hyperref}
\hypersetup{colorlinks=true,urlcolor=blue}

\title{Augmenting Customer Support with an NLP-based Receptionist}

\author{André Barbosa\inst{2}, Alan Godoy\inst{1}}

\address{
    QuintoAndar Inc, São Paulo, Brazil
\nextinstitute
    University of São Paulo, São Paulo, Brazil
    \email{andre.barbosa@ime.usp.br}
    \email{alan.godoy@quintoandar.com.br}
}

\begin{document}

\maketitle

\begin{abstract}
In this paper, we show how a Portuguese BERT model can be combined with structured data in order to deploy a chatbot based on a finite state machine to create a conversational AI system that helps a real-estate company to predict its client’s contact motivation. The model achieves human level results in a dataset that contains 235 unbalanced labels.  Then, we also show its benefits considering the business impact comparing it against classical NLP methods.
\end{abstract}

\section{Introduction}

Effective customer support is an essential piece for any service company. On the one hand, as a business grows, it can hire more agents and create specialized departments focusing on each possible problem a customer may face, increasing the availability of support for clients. On the other hand, growth brings new challenges: knowledge about client issues gets scattered across the company, information regarding processes may not diffuse properly to all agents, clients may have trouble finding the specific department able to help them, to name few. As a real estate company, \href{https://www.quintoandar.com.br/}{QuintoAndar}\footnote{A quick presentation about QuintoAndar and its business model can be found \href{https://forbes.com/sites/angelicamarideoliveira/2019/01/10/quinto-andar-the-brazilian-startup-changing-the-rentals-market-plans-further-disruption/}{here}.} has to deal with a wide range of issues involving individuals as distinct as tenants (and prospective tenants), landlords, agents (realtors, inspectors, photographers, etc) and building administrators. The channel of choice for most contacts is \href{https://www.whatsapp.com/}{WhatsApp}, a messaging app widely adopted in Brazil and other countries. This introduces even more complexity in terms of user identification and routing, as any WhatsApp user can start a conversation with our customer support team simply by sending an open text message.

To improve the routing quality, speed, and automate the collection of all necessary information before a human analyst is allocated to the ticket, our team developed a chatbot system that acts as a receptionist for customer support. This chatbot (presented in section \ref{system-architecture}) uses a state-machine dialog manager to integrate multiple machine learning (ML) based classifiers and business rules to define the messages to be sent and actions to be taken. The implementation of such service brought a lot of positive results to QuintoAndar, allowing the full automation of chat triage in customer support, a task performed previously by a dedicated team, which was reallocated to other departments.

Beyond presenting a chatbot architecture that combines multiple ML models and discussing how we merged textual information with structured data in order to better understand an user's context, the main contribution of this work is to bring data about the results achieved by state-of-the-art models in a real-world application. Much has been discussed about the fantastic results achieved by recent Natural Language Processing (NLP) models for the English language both in academic literature an in industry. How these outcomes transfer to Brazilian Portuguese is not as clear though. We hope that, by sharing our experience, we may shed some light on how Portuguese-speaking companies and research groups may use these technologies outside common benchmark problems.

In summary, the main goal of this paper is to show how we evolve until we achieve human-level results in a real-world application through combining Portuguese BERT with structured metadata from our clients with a complete end-to-end system.

\section{Related Work}

Customer support is a natural candidate to be one of the areas most benefited by the recent boost in NLP research and application~\cite{wan2018how,10.1145/3394486.3403390,10.1145/3219819.3219851}. By having a proper system for customer triage, a company can match its user to the agent most likely to solve their problem and also provide the agent with contextual information --- e.g.: selected user information, procedure suggestion and a list of similar issues --- to make their work more effective. Alexandra DeLucia and Elisabeth Moore~\cite{DeLucia2020AnalyzingHS} used IT support tickets in a high-performance computing laboratory to study automatic categorization and similar ticket recommendation. They combined classical NLP pre-processing, including steps as stop-word removal, stemming and topic modeling, with a random-forest classifier to select the most appropriate label among 93 possible categories and retrieve other similar tickets. A similar work was done by Fotso \emph{et al.}~\cite{Fotso2018AttentionFN}, that created a system to classify client emails in 12 possible categories, using it to select relevant articles to be automatically sent as response. The team combined textual information with data regarding the user's prior interactions with their products, using word-embeddings, a BiLSTM network and an attention-based mechanism to predict the most adequate class.

Rather than applying a single machine learning model to classify the user demand, it's also possible to create a dialog system that interacts with the customer to try solving their problem or to ensure that all relevant information was provided before assigning the ticket to an agent. Such system demands not only a model that extracts information from user messages but also a dialog management module to track the current dialog state and define which action to perform in each moment and also a module to generate the messages sent to the user~\cite{dai2020survey,10.5555/1214993,zhang2020recent}. Multiple approaches can be used for each of these modules. For dialog management they can be as simple as form-filling~\cite{BOBROW1977155}, in which the system has a pre-defined list of information that should be provided for a given task and it will iteratively ask the user to answer each missing item, or get as complex as using a recurrent neural network to infer a latent representation of the current dialog state which is fed to a second network that converts states into policies~\cite{wang-etal-2018-teacher}. 

When considering Brazilian Portuguese, however, there are few works reporting applications of NLP in industrial applications. Finardi \emph{et al.}~\cite{finardi2021bertau} built a BERT language model~\cite{devlin2019bert} to be used for customer support in a large Brazilian bank, evaluating its performance for sentiment analysis, question answering and named entity recognition (NER) using datasets extracted from user interactions with the bank's chatbot. Azevedo \emph{et al.}~\cite{bb2020} created a system to automatically route customer emails to four different boxes using a Support Vector Machine (SVM). Works have also been published regarding applications in law, as \cite{bonifacio20} --- which extensively studies the impacts of fine-tuning of transformer-based language models using legal texts in NER tasks --- and \cite{pont2020} --- that provides an empirical study of word embeddings in legal domain.

On the academic front, however, many works were produced recently related to dialog systems that can be applied for customer support in Portuguese. Santos \emph{et al.}~\cite{santos2020} and Melo and Coheur~\cite{melo2020}, for instance, built retrieval-based conversational agents trained to answer specific data. Carvalho \emph{et al.}~\cite{carvalho2020}, on the other hand, created an LSTM-based common-sense module to augment interactions in a dialog system. A recent significant contribution was also made by Souza \emph{et al.}~\cite{souza2020bertimbau} that made open-source a Brazilian Portuguese pre-trained BERT language model, showing how such mono-lingual model was able to surpass the performance achieved by a multi-lingual BERT for named entity recognition (NER).

\section{Problem Definition and Challenges}

\begin{figure}[ht]
\centering
\includegraphics[width=\textwidth]{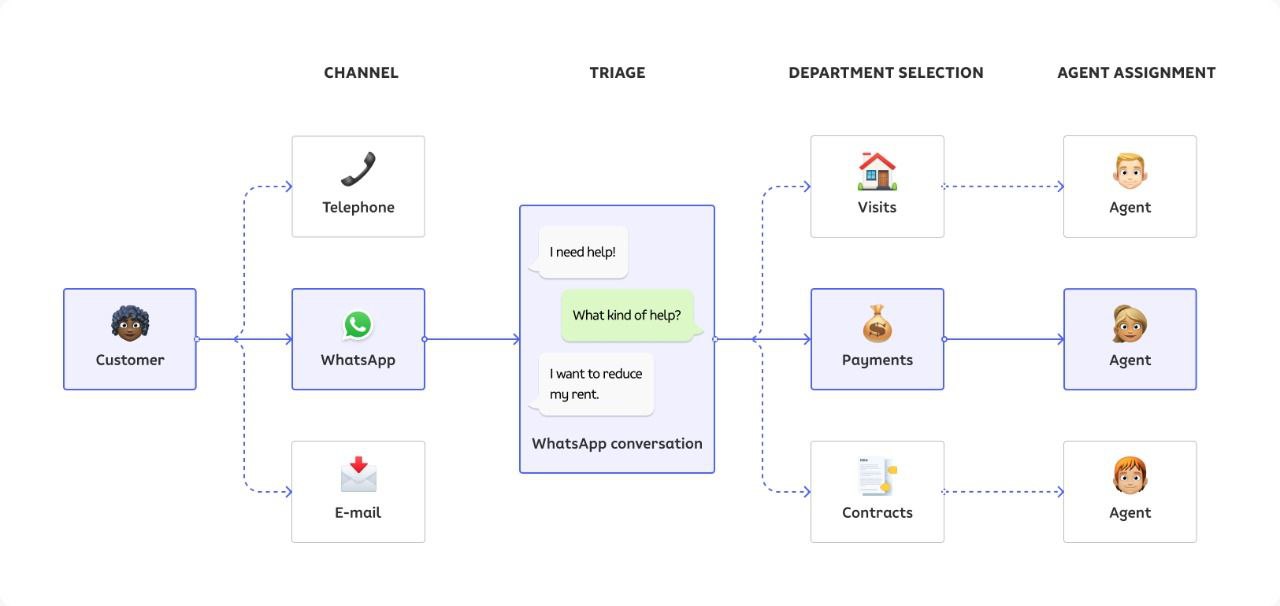}
\caption{A typical flow of customer support at QuintoAndar.}
\label{fig:reception-flow}
\end{figure}

A typical customer support flow at QuintoAndar (Fig.~\ref{fig:reception-flow}) starts with a customer contacting our team through the channel of their choice --- namely, phone, WhatsApp or email/support form. Whichever channel is chosen, the user is required to state what kind of support is needed, an information that is used to define to which department the contact should be directed. After the department is defined, a task allocation system selects among all online agents who should be responsible for handling the customer's issue. The role of such agent is to comprehend the user's demand and get all relevant information to solve it --- if it is simple enough, it can be resolved immediately; if it requires more complex actions, a task is created for the appropriate back-office team.

In a digital real-state company, however, there are hundreds of different types of contacts, going from simple questions about using our site to search for a house to requests for intermediation of complex tenant-owner issues like negotiating a temporary rent reduction. This means that matching a user to the right specialist is easier said than done. On the one hand, presenting a menu with a multitude of customer support departments may be time-consuming and ineffective, as users may not be familiar with each department's responsibilities and are likely to make mistakes when choosing an option. On the other hand, letting the users freely declare what they need has its own drawbacks, such as the possibility that the user's explanation does not have sufficient context regarding their demand and the issue of how to convert that explanation into a decision about the destination department of the contact.

For WhatsApp --- our main support channel and the focus of the present work --- this issue gets trickier, though: the app is ubiquitously used in Brazil as a means of communication with family and friends. Added to this is the fact that Brazilian clients are used to small real estate companies in which commercial interactions are based on a person-to-person relationship. Contacts, thus, may start with a simple \emph{``Hi!''}, with a complete account of the user's issue or with a partial description that refers to past contacts.

Prior to this work, triage and department selection was carried out by a team dedicated solely to verifying that sufficient context was provided, asking users to give more information if necessary, and directing each chat to the appropriate department. Of course, this was a far from ideal situation. A large number of skilled agents were diverted from customer troubleshooting to perform a highly repetitive task that added latency to proper chat routing and stiffened any processes changes that incur in changes in departmental responsibilities, as any such changes would require retraining the triage team.

To sustain company's scalability, our team developed a ``receptionist'' chatbot that should be able to:

\begin{enumerate}
    \item Determine whether the user has provided enough information in their initial message, requesting a better description if deemed necessary.
    \item Combine this textual information with data about the user's relationship with QuintoAndar to predict the reason for the current contact.
    \item Automate the collection of basic data, such as information that helps to identify users who could not be identified based on their phone number and data to help the agent verify the authenticity of the user.
    \item Use business rules together with the predicted contact reason to determine to which department the chat should be directed.
    \item Provide the selected agent with the predicted reason for the contact in order facilitate access to the appropriate guidelines for that service and speed up the mandatory annotation after the conversation is finished.
\end{enumerate}

This was not without its hurdles. First, there were more than 300 standardized possible contact reasons with a substantial gray area between them, ensuring a great deal of noise in labels. Classes were highly unbalanced, with some containing only a handful of chats in a given year, while others classes totaling thousands in a single week. Finally, as a Portuguese-speaking company, there were no open resources available related to customer service, so we could rely only on pre-trained models and our own private datasets to develop the chatbot.

\section{System Architecture}
\label{system-architecture}

Following the dialog-state architecture~\cite{10.5555/1214993}, the bot engine we created has the following basic components:

\begin{itemize}
    \item Dialog memory: a flexible memory to store all relevant information produced by handlers during the conversation, including special field $\mathcal{S}$ that indicates the automaton's current state.
    \item Handlers: processors used by the chatbot to extract relevant information from input messages or make decisions about what to do next.
    \item Finite-state automaton: a recipe that indicates which handler should be called at each state $\mathcal{S}$ and how to use a policy handler results to select the next state.
\end{itemize}

Handlers, in turn, can be of two different types:

\begin{itemize}
    \item Message processing handlers: responsible for extracting information from input messages and storing it on dialog memory. The present work focuses on proposing two machine learning models that act as message processing handlers.
    \item Policy decision handlers: responsible for deciding what the system should do given the information stored in dialog memory. Examples of possible actions are performing some calculations and storing their results in the memory, sending a response to the user or transferring the chat to a certain department. When a policy handler decides to send the user a response, it forwards asks the language generation unit to select a template to use to create the message.
\end{itemize}

Each handler has its own implementation, which may range from simple application of business rules to complex ML models and third-party services. This work makes use of two different NLP-based handlers that will be explained on the next subsections.

\subsection{Context evaluation model}
\label{context-identifier}

A common behavior by our users was to start the conversation by saying a greeting, only stating what she needed after being asked by the support agent. This posed a significant challenge for our receptionist bot, as it would predict contact reasons and select destination departments based on messages that were not at all explanatory. A simple rule of thumb to avoid this problem would be to always discard the first message in a conversation. However, this would hurt the client's experience, which is only worsened by the fact that many customer support contacts are made by already stressed users.

To alleviate this issue, we created a model to predict whether or not a particular message was sufficiently explanatory. Based on such prediction our chatbot may decide to ask the client for more input. A dataset with 5348 chats randomly sampled from QuintoAndar's ticketing system was manually annotated by two customer support agents. They could classify a message as one of the following: \emph{has context}, \emph{no context}, \emph{returning client} and \emph{low value} (a message that is not intended to create a new conversation, as \emph{``Thanks!''}). The final dataset contained 2926 chats labeled as its content has enough context and 2422 labeled as not enough context (\emph{no context} and \emph{low value}; \emph{returning client} messages were discarded), endorsing the relevance of this first model in an efficient reception.

Our goal was to have a simple model that would introduce only a small delay in response time, so we decided to user bag-of-words (BoW) representation with a logistic regression (LR) as classifier. Text pre-processing included accent stripping, uncasing and stop-word removal; no stemming or lemmatization was used. Bag-of-words representation was then computed for 3-grams with a vocabulary of 5000 items. LR hyperparameters ($C$, penalty type and class weight) were selected through Bayesian optimization --- the solver used was `lbfgs` \cite{lbfgs}. The resulting model achieved accuracy of 85\% on the testing split, a value considered good enough for production as most errors in this step could be fixed by human agents after chat allocation.

\subsection{Contact reason prediction model}
\label{contact-reason-model}

The main model for the chatbot is the contact reason prediction model. As explained above, its main purpose is to combine the user's message with tabular information indicating their relationship with QuintoAndar to predict how likely the contact is to be related to each of 306 possible standardized contact reasons (see Table ~\ref{fig:contact-reasons} for some examples).

\begin{table}[ht]
\centering
\begin{tabular}{|l p{0.65\textwidth}|}
\hline
\verb|cr_pg|              & Payments of real-estate brokers. \\
\verb|ft_ag_alteracao|    & Schedule changes for photographers. \\
\verb|iq_pr_reserva|      & Issues related to house reservation by a tenant after a proposal is accepted by an owner. \\
\verb|pp_cm_venda_imovel| & An owner informs that she intends to sell the house she currently rents. \\
\hline
\end{tabular}
\caption{Examples of contact reasons defined by QuintoAndar's process team.}
\label{fig:contact-reasons}
\end{table}

The association of text and tabular data is important as a way to avoid requiring users to explain their whole history to the chatbot. Consider the phrase \emph{``I need to cancel the visit tomorrow.''}. It is a complete yet very ambiguous message: is it related to a photographer that wants to reschedule a photo shoot, to a potential tenant who is not interested on visiting an apartment anymore, or to a real estate agent that will not be able to present some house and wants to leave it to another colleague? Only by having access to the user's relationship with QuintoAndar the model would be able to accurately predict the contact reason without requesting further context. Therefore, we used 66 handcrafted features available from our feature store~\cite{butterfree,uber-michellangelo}. Some examples are the type of the last automatic message sent to the user (and time since it was sent), the contact reason for the last ticket (and time since it was created), whether the user is a registered agent, number of rented houses (as owner) and number of ended contracts (as tenant). These features are pre-processed, performing one-hot encoding on all categorical columns and scaling the numeric ones to have zero mean and unit variance.

We treat both feature groups (textual and tabular) as different modules that are fed into a separate classifier. In a first version (V1 - bag-of-words), we used a simple unigram bag-of-words to extract features from messages. After analyzing this model errors, we noticed that, despite showing good results, it had trouble with synonyms and complex sentences. In a second version of the model (V2 - BERT) we addressed this issue by using for textual feature extraction a representation extracted by a Portuguese version of a BERT model~\cite{devlin2019bert,souza2020bertimbau}.

\begin{figure}[ht]
\centering
\includegraphics[width=\textwidth]{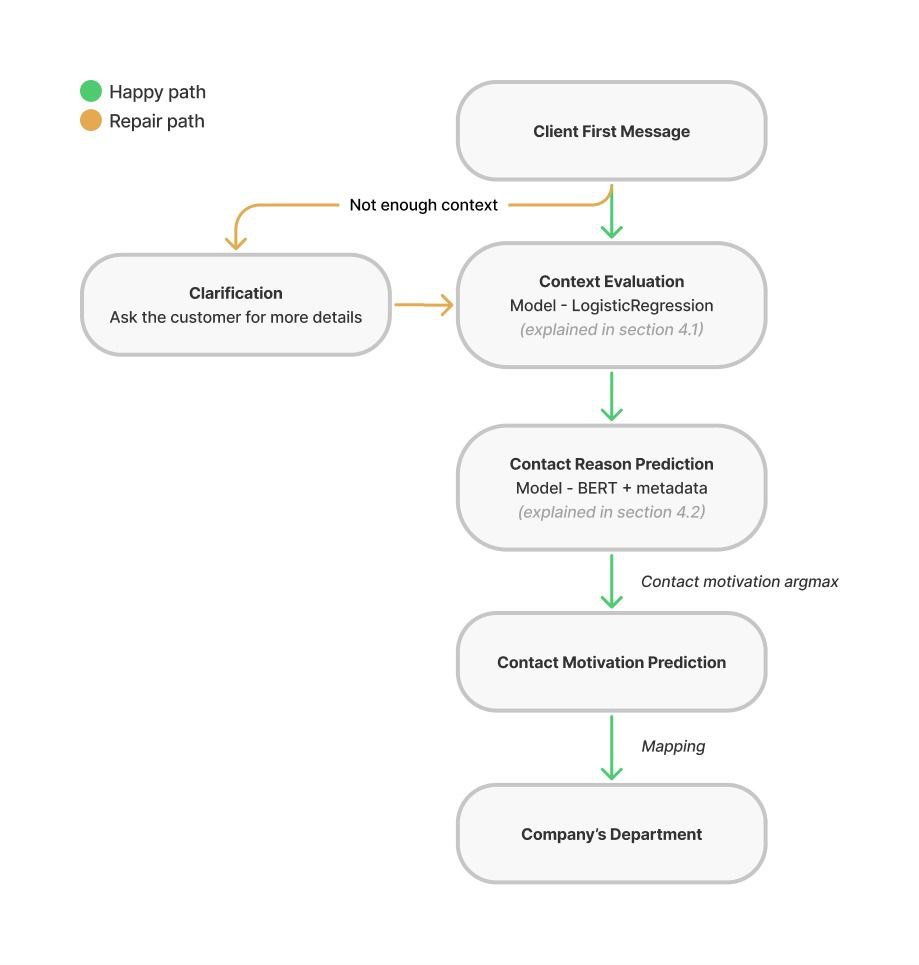}
\caption{An overall picture of system architeture}
\label{fig:system-arch}
\end{figure}

The overall architecture can be summarized in Figure ~\ref{fig:system-arch}.

\section{Experiments}
\label{experiments}

We briefly discuss the results achieved for version 1 and version 2 of contact reason prediction model, taking into consideration different setups as well as analyzing overall model's performance. For more information regarding libraries used and hyperparameters, please refer to supplementary information for more details.

\subsection{Dataset}

The full dataset used to train the contact reason model contains data for 639159 chats between May 2019 and August 2020 manually annotated by support agents, selecting only user messages sent before an agent entered the conversation. We used out-of-time split to partition the between training (511327 chats), validation (63916 chats) and testing (63916 chats) sets. With respect to classes, originally the data contained 306 distinct ones --- after filtering all tickets belonging to classes with less than 50 samples on training set the final dataset contained 235 different labels. To fine-tune BERT on our data, since it is a considerably heavy model, we have used a smaller portion of dataset, with 178578 samples for training, 19843 for validation and 66141 for testing.

\subsection{Metrics}

To analyze the model performance, we have calculated top-1 and top-3 test accuracy both for contact reason and for department\footnote{Department selection is performed by summing predicted probabilities for all contact reasons associated to a given department. The department with highest score is selected.}.

\subsection{V1 - Bag-of-words}

For this first solution approach, we have evaluated three possible classifiers: logistic regression (LR), random forest (RF) and multilayer perceptron classifier (MLP). Hyperparameters were selected through Bayesian optimization with 100 evaluated configurations~\cite{wandb}. Classifiers comparison are presented in Table \ref{table:test-accuracy-classifiers}.

\begin{table}[ht]
\centering
\caption{Test accuracy for contact reason and department. Higher is better.}
\label{table:test-accuracy-classifiers}
\begin{tabular}{|c|c|c|c|c|c|}
\hline
Model & Top-1 accuracy  & Top-3 accuracy  & Top-1 dept. accuracy & Top-3 dept. accuracy \\
\hline
LR    & 42.8\%          & 63.6\%          & 77.8\%               & 84.6\% \\ 
\hline
RF    & 40.7\%          & 61.2\%          & 75.2\%               & 82.7\% \\
\hline
MLP   & \textbf{44.1\%} & \textbf{65.1\%} & \textbf{78.2\%}      & \textbf{85.0\%} \\ 
\hline
\end{tabular}
\end{table}

\subsubsection{V2 - BERT}~As observed in Table \ref{table:test-accuracy-classifiers}, the MLP model obtained the best results according to all metrics. Given the high cost of using a large Transformer-based network and the fact this new version is very similar to V1, we decided to keep the same architecture.

The first step we took was to fine-tune the BERT-Large model made available by Souza \emph{et al.}~\cite{souza2020bertimbau} to predict contact reason. Sentences with more than 64 tokens were truncated. Once this fine-tuned procedure was completed, we tested different methods in order to replace the bag-of-words of V1 with a BERT module:

\begin{itemize}
    \item Use the logits of BERT classification head.
    \item Use the last layer of BERT language model head as suggested by Devlin \emph{et al.}~\cite{devlin2019bert} .
    \item Concatenate last four layers of BERT language model head as suggested by Devlin \emph{et al.}~\cite{devlin2019bert} 
\end{itemize}

The results are shown in Table~\ref{table:bert-comparison}. First of all, it is relevant to notice the large gains achieved by combining both textual and tabular features. Also, we can see that using BERT as textual feature extractor provided better results than using bag-of-words. Considering engineering constraints we have decided to use the logits of BERT classification head + tabular data as version 2 of the contact reason model. 

\begin{table}[ht]
\centering
\caption{Comparison of contact reason accuracy for multiple models.}
\label{table:bert-comparison}
\begin{tabular}{ |c|c| } 
\hline
Model                                       & Top-1 accuracy \\
\hline
BoW alone                                   & 38.14\% \\
\hline
BoW + tabular data (V1)                     & 44.11\% \\
\hline
BERT classifier                             & 45.57\% \\
\hline
BERT classifier logit heads + tabular data  & 53.10\% \\
\hline
\textbf{Last layer LM BERT + tabular data}           & \textbf{53.20}\% \\
\hline
Last 4 layers LM BERT concat + tabular data & 53.05\% \\
\hline
\end{tabular}
\end{table}

\section{Business impacts}

To assess the results in production, we have collected data from a triage human team and for a set of heuristics rules that routed simply based on the last automatic message sent to the user (e.g., if the message was related to a visit, then the user was directed to the visits department). The business metric that we decided to follow to compare was the transference rate --- \emph{i.e.}, the rate in which a chat already routed to a given support department should be transferred to another department.

We have run tests in production comparing human triage and both chatbot versions. As a security measure to avoid deterioration of user experience, we only routed automatically the 80\% of tickets with highest department score leaving all low-confidence tickets to human triage. As the results were good enough (i.e. similar to human performance), we expanded it to 100\% to out clients base. Results are presented in Table~\ref{table:production-results}.

\begin{table}[ht]
\centering
\caption{Production results. Data for human triage and heuristic rules refer were collected prior the tests. Lower the better for both columns.}
\label{table:production-results}
\begin{tabular}{ |c|c|c| } 
\hline
Model                    & Transf. rate & Avg. msg. per ticket \\
\hline
Human triage             & 12.8\%       & 18.2 \\
\hline
Heuristics rules         & 18.3\%       & 11.2 \\
\hline
V1 - Bag-of-words (80\%) & 13.9\%       & 13.2 \\
\hline
V2 - BERT (80\%)         & 10.3\%       & 13.7 \\
\hline
V2 - BERT (100\%)        & 13.2\%       & 14.2 \\
\hline
\end{tabular}
\end{table}

Considering these results, we can easily say that the BERT embedding combined with tabular data achieved human-level performance. The number of message exchanges until the conversation was ended (avg. message per ticket)  was also substantially smaller for clients routed by the chatbot than for those routed by humans. An hypothesis regarding the reduction in message numbers is the fact that by using tabular features the model has access to a large amount of information not easily consumed by humans.

Such results in a real-world application using business metrics endorses the potential of modern NLP techniques for Brazilian Portuguese. We hope that results like these help companies and governments to see that it's now feasible to go much beyond the widespread rigid conversational interfaces based on buttons or simple keywords. By creating applications based on recent breakthroughs, we can make services both more efficient and accessible.

\section*{Acknowledgments}
We would like to thanks Muriel Dias for providing beautiful images that help in our paper storytelling, as well as Marco Antonio Rocha Vinha for reviewing some components of system architectures that were showed in section 4.
\bibliographystyle{sbc}
\bibliography{stil-complete}

\begin{thebibliography}{}

\bibitem[Azevedo et~al. 2020]{bb2020}
Azevedo, R. F.~d., Rodrigues Pereira~de Araujo, R., Guimar{\~a}es~Ara{\'u}jo,
  R., Moreira~Bittencourt, R., Ferreira Alves da Silva, R., de~Melo
  Vaz~Nogueira, G., Marques~Franca, T., Otharan~Nunes, J., Ralff~da Silva, K.,
  and Regiane Cunha~de Oliveira, E. (2020).
\newblock Screening of email box in portuguese with svm at banco do brasil.
\newblock In Quaresma, P., Vieira, R., Alu{\'i}sio, S., Moniz, H., Batista, F.,
  and Gon{\c{c}}alves, T., editors, {\em Computational Processing of the
  Portuguese Language}, pages 153--163, Cham. Springer International
  Publishing.

\bibitem[Biewald 2020]{wandb}
Biewald, L. (2020).
\newblock Experiment tracking with weights and biases.
\newblock Software available from wandb.com.

\bibitem[Bobrow et~al. 1977]{BOBROW1977155}
Bobrow, D.~G., Kaplan, R.~M., Kay, M., Norman, D.~A., Thompson, H., and
  Winograd, T. (1977).
\newblock {GUS}, a frame-driven dialog system.
\newblock {\em Artificial Intelligence}, 8(2):155--173.

\bibitem[Bonifacio et~al. 2020]{bonifacio20}
Bonifacio, L.~H., Vilela, P.~A., Lobato, G.~R., and Fernandes, E.~R. (2020).
\newblock A study on the impact of intradomain finetuning of deep language
  models for legal named entity recognition in portuguese.
\newblock In Cerri, R. and Prati, R.~C., editors, {\em Intelligent Systems},
  pages 648--662, Cham. Springer International Publishing.

\bibitem[Carvalho et~al. 2020]{carvalho2020}
Carvalho, C.~S., Pinheiro, V.~C., and Freire, L. (2020).
\newblock A deep learning model of common sense knowledge for augmenting
  natural language processing tasks in portuguese language.
\newblock In Quaresma, P., Vieira, R., Alu{\'i}sio, S., Moniz, H., Batista, F.,
  and Gon{\c{c}}alves, T., editors, {\em Computational Processing of the
  Portuguese Language}, pages 303--312, Cham. Springer International
  Publishing.

\bibitem[Dai et~al. 2020]{dai2020survey}
Dai, Y., Yu, H., Jiang, Y., Tang, C., Li, Y., and Sun, J. (2020).
\newblock A survey on dialog management: Recent advances and challenges.

\bibitem[Dal~Pont et~al. 2020]{pont2020}
Dal~Pont, T.~R., Sabo, I.~C., H{\"u}bner, J.~F., and Rover, A.~J. (2020).
\newblock Impact of text specificity and size on word embeddings performance:
  An empirical evaluation in brazilian legal domain.
\newblock In Cerri, R. and Prati, R.~C., editors, {\em Intelligent Systems},
  pages 521--535, Cham. Springer International Publishing.

\bibitem[DeLucia and Moore 2020]{DeLucia2020AnalyzingHS}
DeLucia, A. and Moore, E. (2020).
\newblock Analyzing hpc support tickets: Experience and recommendations.
\newblock {\em ArXiv}, abs/2010.04321.

\bibitem[Devlin et~al. 2019]{devlin2019bert}
Devlin, J., Chang, M.-W., Lee, K., and Toutanova, K. (2019).
\newblock {BERT}: Pre-training of deep bidirectional transformers for language
  understanding.

\bibitem[Finardi et~al. 2021]{finardi2021bertau}
Finardi, P., Viegas, J.~D., Ferreira, G.~T., Mansano, A.~F., and Caridá, V.~F.
  (2021).
\newblock Berta\'u: Ita\'u bert for digital customer service.

\bibitem[Fotso et~al. 2018]{Fotso2018AttentionFN}
Fotso, S., Spanoudes, P., Ponedel, B.~C., Reynoso, B., and Ko, J. (2018).
\newblock Attention fusion networks: Combining behavior and e-mail content to
  improve customer support.
\newblock {\em ArXiv}, abs/1811.03169.

\bibitem[Hermann and Balso 2017]{uber-michellangelo}
Hermann, J. and Balso, M.~D. (2017).
\newblock Meet michelangelo: Uber’s machine learning platform.

\bibitem[Jurafsky and Martin 2021]{10.5555/1214993}
Jurafsky, D. and Martin, J.~H. (2021).
\newblock {\em Speech and Language Processing (3rd Edition)}.
\newblock Prentice-Hall, Inc., USA.

\bibitem[Liu et~al. 2020]{10.1145/3394486.3403390}
Liu, C., Jiang, J., Xiong, C., Yang, Y., and Ye, J. (2020).
\newblock Towards building an intelligent chatbot for customer service:
  Learning to respond at the appropriate time.
\newblock In {\em Proceedings of the 26th ACM SIGKDD International Conference
  on Knowledge Discovery \& Data Mining}, KDD 20, page 3377–3385, New York,
  NY, USA. Association for Computing Machinery.

\bibitem[Marques 2020]{butterfree}
Marques, A. (2020).
\newblock Butterfree: A spark-based framework for feature store building.

\bibitem[Melo and Coheur 2020]{melo2020}
Melo, G. and Coheur, L. (2020).
\newblock Towards a conversational agent with ``character''.
\newblock In Quaresma, P., Vieira, R., Alu{\'i}sio, S., Moniz, H., Batista, F.,
  and Gon{\c{c}}alves, T., editors, {\em Computational Processing of the
  Portuguese Language}, pages 420--424, Cham. Springer International
  Publishing.

\bibitem[Molino et~al. 2018]{10.1145/3219819.3219851}
Molino, P., Zheng, H., and Wang, Y.-C. (2018).
\newblock {COTA}: Improving the speed and accuracy of customer support through
  ranking and deep networks.
\newblock In {\em Proceedings of the 24th ACM SIGKDD International Conference
  on Knowledge Discovery \& Data Mining}, KDD 18, page 586–595, New York, NY,
  USA. Association for Computing Machinery.

\bibitem[Santos et~al. 2020]{santos2020}
Santos, J., Alves, A., and Gon{\c{c}}alo~Oliveira, H. (2020).
\newblock Leveraging on semantic textual similarity for developing a portuguese
  dialogue system.
\newblock In Quaresma, P., Vieira, R., Alu{\'i}sio, S., Moniz, H., Batista, F.,
  and Gon{\c{c}}alves, T., editors, {\em Computational Processing of the
  Portuguese Language}, pages 131--142, Cham. Springer International
  Publishing.

\bibitem[Souza et~al. 2020]{souza2020bertimbau}
Souza, F., Nogueira, R., and Lotufo, R. (2020).
\newblock {BERTimbau}: pretrained {BERT} models for {Brazilian Portuguese}.
\newblock In {\em 9th Brazilian Conference on Intelligent Systems, {BRACIS},
  Rio Grande do Sul, Brazil, October 20-23}.

\bibitem[Wan and Chen 2018]{wan2018how}
Wan, M. and Chen, X. (2018).
\newblock Beyond "how may i help you?": Assisting customer service agents with
  proactive responses.

\bibitem[Wang et~al. 2018]{wang-etal-2018-teacher}
Wang, W., Zhang, J., Zhang, H., Hwang, M.-Y., Zong, C., and Li, Z. (2018).
\newblock A teacher-student framework for maintainable dialog manager.
\newblock In {\em Proceedings of the 2018 Conference on Empirical Methods in
  Natural Language Processing}, pages 3803--3812, Brussels, Belgium.
  Association for Computational Linguistics.

\bibitem[Zhang et~al. 2020]{zhang2020recent}
Zhang, Z., Takanobu, R., Zhu, Q., Huang, M., and Zhu, X. (2020).
\newblock Recent advances and challenges in task-oriented dialog system.

\bibitem[Zhu et~al. 2011]{lbfgs}
Zhu, C., Byrd, R., Nocedal, J., and Morales, J.~L. (2011).
\newblock {L-BFGS-B} --- software for large-scale bound-constrained
  optimization.

\end{thebibliography}

\end{document}